\documentclass{article}

\usepackage{natbib}
\setcitestyle{numbers,square}

\usepackage[preprint]{neurips_2023}

\pdfoutput=1
\usepackage[utf8]{inputenc} 
\usepackage[T1]{fontenc}    
\usepackage[hidelinks]{hyperref}       
\usepackage{url}            
\usepackage{booktabs}       
\usepackage{amsfonts}       
\usepackage{nicefrac}       
\usepackage{microtype}      
\usepackage{xcolor}         
\usepackage{floatrow}
\usepackage{graphicx}
\usepackage{physics}
\usepackage{hyperref}
\usepackage[capitalize]{cleveref}
\usepackage{chngcntr}
\newcommand{\setParDis}{\setlength {\parskip} {0.0cm} }
\newcommand{\setParDef}{\setlength {\parskip} {0pt} }

\newcommand{\myparagraph}[1]{\vspace{1pt} \noindent \textbf{#1}}

\newcommand{\ours}{\textit{DreamIdentity}}

\newcommand{\ie}{\textit{i}.\textit{e}.}
\newcommand{\eg}{\textit{e}.\textit{g}.} 

\newcommand{\editb}{\textit{editability}}

\newcommand{\Imetric}{\textit{ID-preservation}}

\newcommand{\task}{\textit{identity re-contextualization}}

\usepackage{float}
\usepackage{color}

\title{\ours: 
Improved Editability for Efficient Face-identity Preserved Image Generation}

\author{
    Zhuowei Chen\textsuperscript{\rm 1,2}\quad 
    Shancheng Fang\textsuperscript{\rm 2}\quad  
    Wei Liu\textsuperscript{\rm 2} \quad Qian He\textsuperscript{\rm 2} \quad
    Mengqi Huang\textsuperscript{\rm 1}\\ 
    \textbf{Yongdong Zhang\textsuperscript{\rm 1}\quad
    Zhendong Mao\textsuperscript{\rm 1}\thanks{Corresponding authors} }\\
    \textsuperscript{\rm 1}~University of Science and Technology of China \quad 
    \textsuperscript{\rm 2}~ByteDance Inc.\quad  \\
    \{chenzw01, huangmq\}@mail.ustc.edu.cn \quad \{zdmao, zyd73\}@ustc.edu.cn \\
    \{fangshancheng.lh, liuwei.jikun, heqian\}@bytedance.com\\
\textcolor{red}{\url{https://dreamidentity.github.io/}}
}

\begin{document}

\maketitle
\begin{abstract}\label{sec:abstract}
While large-scale pre-trained text-to-image models can synthesize diverse and high-quality human-centric images, an intractable problem is how to preserve the face identity for conditioned face images. Existing methods either require time-consuming optimization for each face-identity or learning an efficient encoder at the cost of harming the editability of models. In this work, we present an optimization-free method for each face identity, meanwhile keeping the editability for text-to-image models. Specifically, we propose a novel face-identity encoder to learn an accurate representation of human faces, which applies multi-scale face features followed by a multi-embedding projector to directly generate the pseudo words in the text embedding space. Besides, we propose self-augmented editability learning to enhance the editability of models, which is achieved by constructing paired generated face and edited face images using celebrity names, aiming at transferring mature ability of off-the-shelf text-to-image models in celebrity faces to unseen faces. Extensive experiments show that our methods can generate identity-preserved images under different scenes at a much faster speed.
\end{abstract}
\section{Introduction}\label{sec:intro}

Diffusion-based large-scale text-to-image (T2I) models \cite{ramesh2022hierarchical, saharia2022photorealistic, rombach2022high} have revolutionized the field of visual content creation recently. With the help of these T2I models, it is now possible to create vivid and expressive human-centric images easily. An exciting application of these models is that, given a specific person's face in our personal life (our family members, friends, etc.), they potentially can create different scenes associated with this identity using natural language descriptions.

\begin{figure}[t!]
  \centering
  \includegraphics[width=\linewidth]{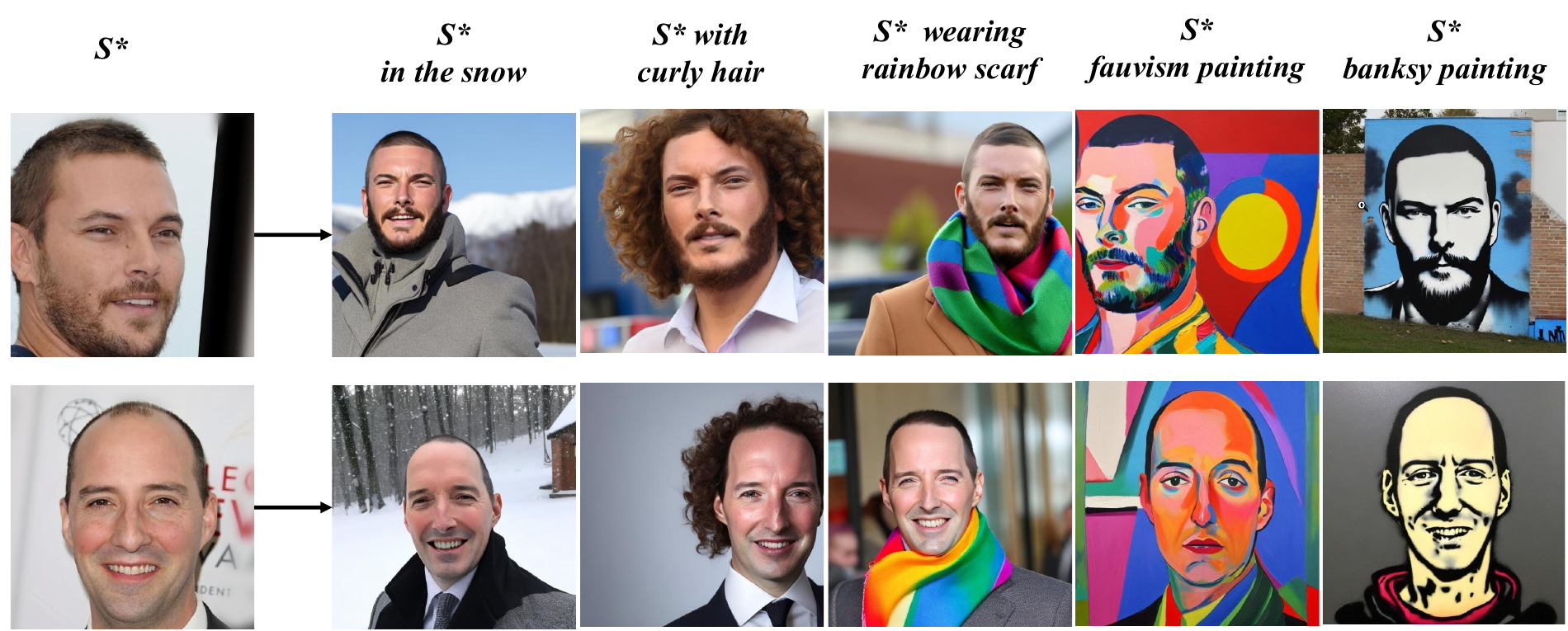}
  \caption{Given only one facial image, \ours \ can efficiently generate countless identity-preserved and  text-coherent images in different context without any test-time optimization.}
  \label{fig:teaser}
\end{figure}

Deviated from the standard T2I task, as shown in Fig.\ref{fig:teaser}, the task \task \ requires the model to have the ability to preserve input face identity (\ie, ID-preservation) while adhering to textual prompts. A feasible solution is to personalize a pre-trained T2I model \cite{gal2022image, ruiz2022dreambooth, kumari2022multi} for each face identity, which involves learning to associate a unique word with the identity by optimizing its word embedding \cite{gal2022image} or tuning the model parameters \cite{ruiz2022dreambooth, kumari2022multi}. However, these optimization-based methods are highly inefficient due to the per-identity optimization. Afterwards, several optimization-free methods \cite{wei2023elite, shi2023instantbooth, ma2023unified} propose to directly map the image features extracted from a pre-trained image encoder (typically, CLIP) into a word embedding, eliminating the cumbersome per-identity optimization but at the cost of degraded ID-preservation. Consequently, these methods necessitate either fine-tuning the parameters of the pre-trained T2I model \cite{gal2023designing} or adjusting original structure for injecting additional grid image features, as a result of bringing the risk of compromising original T2I model's editing capabilities. In a word, all concurrent optimization-free works struggle to preserve identity while remaining the model's editability.

We argue that the above problem of existing optimization-free works arises from  two issues, \ie, (1) the inaccurate identity feature representation and (2) the inconsistency objective between the  training and testing. On one side, the common encoder (\ie, CLIP \cite{radford2021learning}) used by concurrent works \cite{wei2023elite, shi2023instantbooth, ma2023unified, gal2023designing} is unsuitable for identity re-contextualization task as evident by the fact that the current best CLIP model is still much worse than the face recognition model on top-1 face identification accuracy ($80.95\%$ vs $87.61\%$ \cite{bhat2023face}). Additionally, the last layer feature of CLIP struggles to preserve the identity information since it primarily contains high-level semantics, lacking detailed facial descriptions. On another side, all concurrent works solely adopt the vanilla reconstruction objective to learn the word embedding, which hurts the editability for the input face.

In this work, we introduce a novel optimization-free framework (dubbed as \emph{DreamIdentity}) with \emph{accurate identity representation} and \emph{consistent training/inference objective} to deal with the above challenge on identity-preservation and editability. To be specific, for the accurate identity representation,  we design a novel  Multi-word Multi-scale  ID encoder ($M^2$ ID encoder) in the architecture of Vision Transformer \cite{dosovitskiy2020image}, which is pre-trained on a large-scale face dataset and projects multi-scale features into multi-word embeddings.
For the consistent training/inference objective, we propose a novel Self-Augmented Editability Learning to take the editing task into the training phase, which utilizes the T2I model itself to construct a self-augmented dataset by generating  celebrity faces along with various target edited celebrity images. This dataset is then employed to train the $M^2$ ID encoder to enhance the model's editability.

Our contributions in this work are as follows:

\textbf{Conceptually,} we point out current optimization-free methods fail for both ID-preservation and high editability   since their inaccurate representation and inconsistency training/inference objective.

\textbf{Technically,} (1) for accurate representation, we propose $M^2$ ID Encoder, an ID-aware multi-scale feature with multi-embedding projection. (2) For consistent training/inference objective, we introduce self-augmented editability learning to generate a high-quality dataset by the base T2I model itself for editing.

\textbf{Experimentally,} extensive experiments demonstrate the superiority of our methods, which can efficiently achieve identity-preservation while enabling flexible text-guided editing, 
\ie, identity re-contextualization.

\vspace{-0.3cm}
\section{Related Work}\label{sec:related}
\vspace{-0.2cm}
\subsection{Text-to-image Generation}
Text-to-Image generation aims to generate realistic and semantically consistent images with natural language descriptions. Early works mainly adopted GAN \cite{goodfellow2014generative} as the foundational generative model for this task. Various works have been proposed \cite{zhang2021cross, zhu2019dm, xu2018attngan, zhang2017stackgan, zhang2018photographic, liang2020cpgan, cheng2020rifegan, ruan2021dae, tao2020df, li2019controllable, huang2022dse} with well-designed text representation, elegant text-image interaction, and effective loss function. However, GAN-based models often suffer from training instability and model collapse, making it hard to be trained on large-scale datasets \cite{brock2018large, kang2023scaling, schuhmann2021laion}. Witnessed by the scalability of large language models \cite{radford2019language}, autoregressive methods like DALL-E and Parti \cite{yu2022scaling, ramesh2021zero} where the images are quantized into discrete tokens are scaled to learn more general text-to-image generation.
More recently, Diffusion Models, such as GLIDE \cite{nichol2021glide}, Imagen \cite{saharia2022photorealistic},  DALL-E 2 \cite{ramesh2022hierarchical}, LDM \cite{rombach2022high}  have demonstrated the ability on generating unprecedentedly high-quality and diverse images.  However, it remains infeasible to generate a specified identity within the context described by the text. However, generating a specified face/person identity within the context described by the text using the text-to-image model alone remains infeasible.

\begin{figure}[tb]
  \centering
  \includegraphics[width=\linewidth]{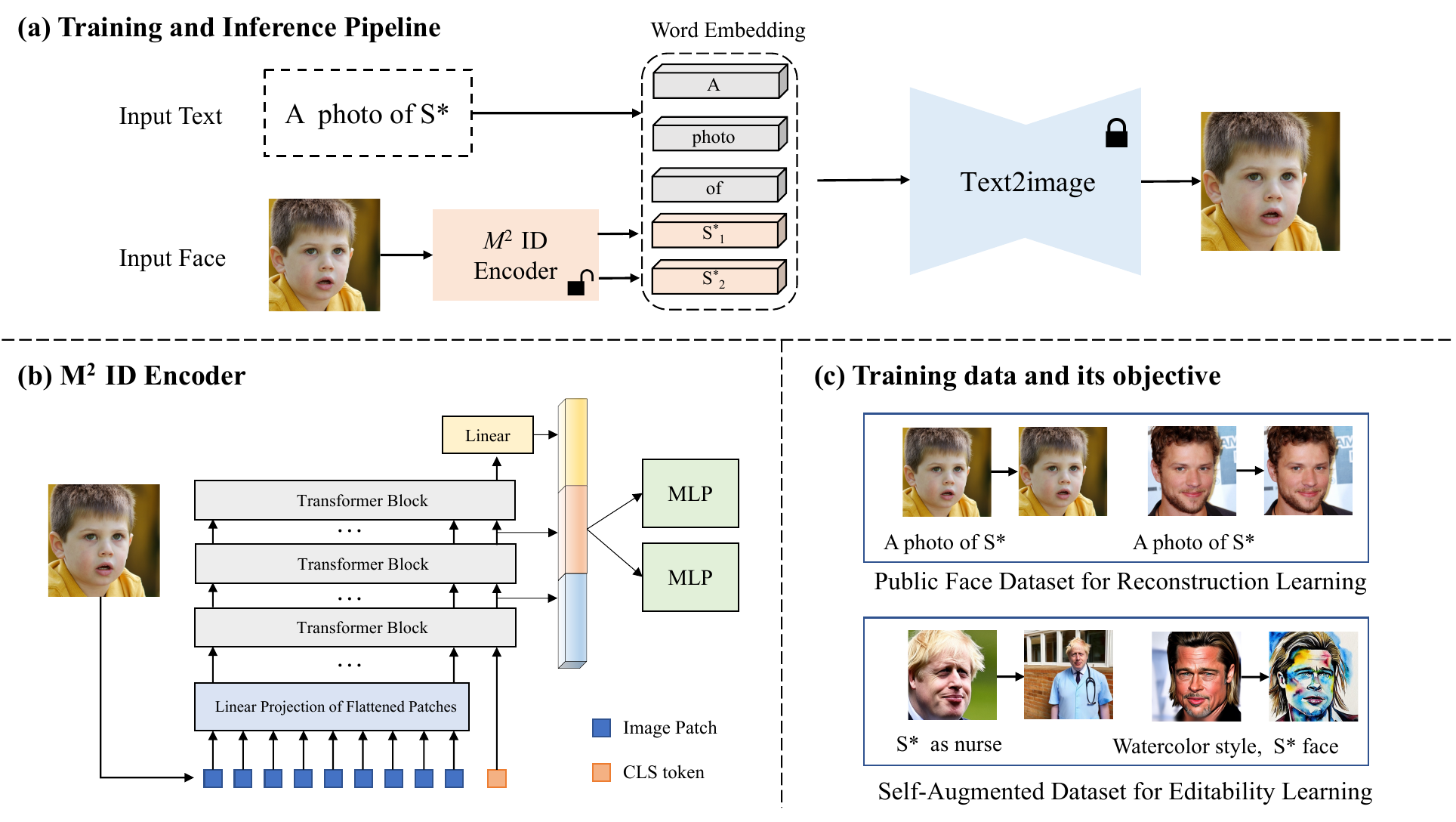}
  \caption{Overview of the proposed \ours: (a) The training and inference pipeline. The input face image is first encoded into multi-word embeddings (denoted by $S^*$) by our proposed $M^2$ ID encoder. Then $S^*$ are associated with the text input to generate face-identity preserved image in the text-aligned scene. (b) The architecture of $M^2$ ID encoder, where a ViT-based face identity encoder is adopted as the backbone and the extracted multi-scale features are projected to multi-word embedding. (c) The composition of the training data and its objectives. The training data consists of a public face dataset for reconstruction and a self-augmented dataset for  editability learning. }
  \label{fig:pipeline} 
\end{figure}
\subsection{Personalized Image Synthesis for Face Identity Control}
Recently personalization methods \cite{gal2022image, ruiz2022dreambooth, kumari2022multi}  have shown promising results on customized concept generation. We can apply these methods to our tasks when the concept is a specified face identity. Textual Inversion \cite{gal2022image} optimizes a new word embedding to represent the given specific concept. \cite{ruiz2022dreambooth} \cite{kumari2022multi} associate the concept with a rare word embedding by fine-tuning part or all parameters in the generator.
However, the requirement for multiple images to specify a concept, coupled with the time-consuming optimization (requiring at least several minutes), limits wider application. Our work present an optimization-free identity encoder that directly encodes a face identity as the word embedding given only one image.

Similar to our goal, there are some concurrent works utilize an embedding encoder for efficient personalized image synthesis. Specifically, ELITE \cite{wei2023elite}, UMM-Diffusion \cite{ma2023unified} and InstantBooth \cite{shi2023instantbooth} encode a common object as a word embedding with the last layer feature from the CLIP encoder. Additionally, ELITE and InstantBooth augment finer details with a local mapping network. Our work differs in several aspects:  1) At the encoder level:  We design a dedicated ID encoder for accurate face encoding with multi-scale features along with multiple word embeddings mapping, whereas the concurrent works use a last layer feature to predict a single word embedding with a common object encoder (CLIP). 2) We propose a self-augmented editability learning method to improve the editability instead of training encoder solely under the reconstruction objective.

\section{Methods}\label{sec:methods}

Given a single facial image of an individual, our objective is to endow the pre-trained T2I model with the ability to efficiently re-contextualize this unique identity under various textual prompts. These prompts may include variations in clothing, accessories, styles, or backgrounds.

The overall framework is shown in Fig.\ref{fig:pipeline}, given a pre-trained T2I model, 
to achieve fast and identity-preserved image generation, we first directly encode the target identity into the word embedding  space (represented as the pseudo word $S*$) with the proposed $M^2$ ID encoder. Afterward,
$S*$ is integrated with the input template prompt for 
generating the text-guided image. To empower the editability for the target identity, a novel \emph{self-augmented editability learning} is further introduced to train the $M^2$ ID encoder with the editability objective.

In the following parts, we first briefly introduce the pre-trained diffusion-based text-to-image model used in our work, then describe our proposed  $M^2$ ID encoder and self-augmented editability learning in detail, respectively.

\subsection{Preliminary}
In this work, we adopt the open-sourced Stable Diffusion 2.1-base (SD) as our text-to-image model, which has been trained on billions of images and shows amazing image generation quality and prompt understanding. 

SD is a kind of Latent Diffusion Model (LDM) \cite{rombach2022high}. LDM firstly represents the input image $x$ in a lower resolution latent space $z$ via a Variational Auto-Encoder (VAE) \cite{kingma2013auto}. Then a text-conditioned diffusion model is trained to generate the latent code of the target image from text input $c$. The loss function of this diffusion model can be formulated as:
\begin{equation}
    \mathcal{L}_{diffusion} = \mathbb{E}_{\epsilon,z,c,t}[\lVert{\epsilon - \epsilon_{\theta}(z_t,c,t)}\rVert_2^2],
\end{equation}
where $\epsilon_{\theta}$ is the noise predicted by the model with learnable parameters $\theta$, $\epsilon$ is noise sampled from standard normal distribution, $t$ is the time step, and $z_t$ is noisy latent at the time step $t$.

During inference, the image is generated by two stages: latent code is first generated by the diffusion model, then the decoder is employed to map the latent code to image space. 

\subsection{$M^{2}$ ID Encoder}

To accurately represent the input face identity, we propose a novel Multi-word Multi-scale embedding ID encoder ($M^2$ ID encoder) for an accurate mapping, which is achieved by the multi-scale ID features extracted from a dedicated backbone then followed by multiple word embedding projection.

\myparagraph{Backbone.} We argue that an accurate representation of the face identity is crucial, while common image encoder CLIP (which is adopted by \emph{all} existing works) fails for that purpose since it can not capture the identity feature as accurately as the face ID encoder which has been trained for face identification tasks on the large-scale face dataset. As \cite{bhat2023face} shows, the current best CLIP VIT-L/14 model is still much worse than the face recognition model on top-1 face identification accuracy ($80.95\%$ vs $87.61\%$). Therefore, we employ a ViT backbone \cite{dosovitskiy2020image} pre-trained on a large-scale face recognition dataset to faithfully extract ID-aware features for input face.

\myparagraph{Multi-scale Feature.}  However, naively mapping the final layer's output identity vector $v_{final}$ could only bring sub-optimal identity preservation. The reason lies in that $v_{final}$ mainly contains the high-level semantics which is suitable for discriminative tasks (\eg, face identification) rather than generative tasks. For example, the same identity with different expressions should share similar representation under the face recognition training loss, while the generation requests more detailed information like facial expressions. Hence, only mapping the last layer representation could become an information bottleneck for the image generation task. To deal with the above problem, we propose to utilize multi-scale features from the face encoder to represent an identity more faithfully. Specifically, the identity vector is augmented by four CLS embeddings ($v_3$, $v_6$, $v_{12}$, $v_{12}$) from the 3rd, 6th, 9th, and 12th layer, respectively. Formally, the multi-scale feature from the ID encoder is depicted as follows:
\begin{equation}
V = [v_3, v_6, v_9, v_{12}, v_{final}].
\end{equation}

\myparagraph{Multi-word Embeddings.} The multi-scale feature is further projected into the word embedding domain. To maintain the original large-scale T2I model's generalization and editability, we leave all its parameters and structure unchanged. As a result, it raises the problem that a single word embedding is hard to faithfully represent the face's identity. Therefore, we further propose a multi-word projection mechanism to represent a face with multi-word embedding:
\begin{equation}
\begin{aligned}
s_{i} = MLP_i(V), \text{for } i = 1, ..., k,
\end{aligned}
\end{equation}
where $k$ is the number of embeddings . Experimentally, we set $k=2$ as depicted in Fig.\ref{fig:pipeline}.  Following \cite{gal2023designing}, $l_2$ regularization is further adapted to constrain the output embedding:

\begin{equation}
    \mathcal{L}_{reg} = \sum_{i=1}^k{\lVert{s_{i}}\rVert}.
\end{equation}

Benefiting from the above-dedicated ID feature, we can facilitate highly identity-preservation control in the embedding space only, without sacrificing pre-trained T2I model's editability caused by feature injection. 

\subsection{Self-Augmented Editability  
 Learning}
Current efficient methods are trained under the reconstruction paradigm, which is given an input face image $I$, the objective to learn a unique word $S*$ such that the $S*$ can reconstruct $I$. However, in real-world applications, we wish to generate a set of new images, such as "watercolor style of $S*$ face", "$S*$ as a police". As a result, there exists a huge inconsistency between training and testing. We hope we can rely on the inductive bias in the word embedding space to achieve editability, but in reality, as Fig.\ref{fig:exp_self_aug} shows, the generated image doesn't always follow the text prompt if we only train encoder under the reconstruction objective. 

To deal with the inconsistency between training and testing, in this paper, we propose a novel \emph{self-augmented editability learning} to take the editing task into the training phase. However, collecting such pair data for the editing task is difficult. Experimentally,  we notice that the current state-of-the-art general text-to-image models can generate celebrity (\eg, Boris Johnson, Emma Watson) in different contexts with good identity preservation and text-coherence. With this insight, The \emph{self-augmented editability learning} utilizes the pre-trained model itself to construct a self-augmented dataset by generating various celebrity faces along with the target edited celebrity images, which will be used to train the $M^2$ ID encoder with the editability objective. Formally, the construction of the dataset includes the following four steps:

\myparagraph {Step 1: Celebrity List Generation.} Firstly we collect a candidate celebrity list. The large language model (\ie, ChatGPT) is used to generate the most famous 400 names in four fields (\ie, sports players, singers, actors, and politicians). After filtering duplicate ones, we finally get 1015 celebrity names.

\myparagraph {Step 2: Celebrity Face Generation.} We propose to use generated face images rather than real images because the model has its own understanding of celebrity. Specifically, the celebrities who appeared less frequently in the Stable Diffusion training dataset are not very similar to the real person while these generated faces maintain a high level of identity resemblance. We use the prompt template "<celebrity-name> face, looking at the camera" to produce the source images, then followed by face crop and alignment operation to get face-only images. A face-only image is kept if its short size is larger than 128 pixels. 

\myparagraph {Step 3: Edit Prompts and Edited Images Generation.} We manually design a variety of prompts that contain images of celebrities in different jobs, styles, and accessories (\eg, "<celebrity-name> as a chef", "oil painting style, <celebrity-name> face"). Then these prompts are transformed to images by the T2I model as edited images, and the <celebrity-name> in prompts is replaced by the pseudo word $S^*$ as Editing Prompts.

\myparagraph {Step 4: Data Cleaning.} After the above procedures, we can now get the initial self-augmented dataset consisting of a set of triplets, <identity face, editing prompt, edited images>. Due to the instability of the current diffusion model, the edited images don't always follow the edit instructions. 
Therefore, we need to filter out the noise data in the self-augmented dataset. We employ ID Loss and CLIP score which reflect identity similarity and text-image consistency as the metrics to rank the edited images for every prompt, then the top $25\%$ triplets at kept as the final training set. 

Finally, we construct a high-quality self-augmented dataset from the pre-trained T2I model itself, which is then used for edit-oriented training.

\subsection{Training}
We combine the FFHQ \cite{karras2019style} and the self-augmented dataset to train our proposed $M^2$ ID encoder. The total loss consists of noise prediction loss of diffusion and the embedding regularization loss:  
\begin{equation}
\mathcal{L}_{total} = \mathcal{L}_{diffusion} + \lambda \mathcal{L}_{reg} ,
\end{equation} 
where $\lambda$ is the embedding regularization weight.

\begin{table*}[t]
\centering
  \caption{Quantitative comparisons with the optimization-based and efficient methods. Encoding time means the time cost to obtain the unique/pseudo embedding. Our method achieves optimal results in terms of text-alignment, face similarity, and encoding time.}
  \label{tab:main_result}
  \begin{tabular}{cccc}
    \toprule
    Methods & Text-alignment $\uparrow$ & Face similarity $\uparrow$ & Encoding Time $\downarrow$ \\
    \midrule
    Textual Inversion \cite{gal2022image} & 0.213 & 0.326 & 20 min \\ 
    Dreambooth \cite{ruiz2022dreambooth} & 0.217 & 0.425 & 4 min  \\ 
    E4T \cite{gal2023designing} & 0.220 & 0.420 & 20 s \\ 
    Elite \cite{wei2023elite} & 0.196 & 0.450 & 0.05 s\\ 
    \midrule
    Ours & \textbf{0.228} & \textbf{0.467} & \textbf{0.04 s}\\
    \bottomrule
  \end{tabular}
\end{table*}

\section{Experiments}\label{sec:exp}

\subsection{Experiment Settings}
\myparagraph{Dataset.} Our experiments are conducted on the widely used FFHQ dataset \cite{karras2019style}, which contains $70000$ high-resolution human face images. We resize the images to 512x512 for training. The test set consists of 100 faces from \cite{liu2015faceattributes}. We make certain that there is no intersection between the test set and the self-augmented celebrity set to maintain the integrity of the experiment.

\myparagraph{Metrics.} We evaluate our method on Text-alignment and Face-similarity.  Text-alignment is used to indicate whether the generated image reflects editing prompts, which is calculated by the cosine distance in the CLIP text-image embedding space.  Face-similarity is used to measure whether the face ID is preserved. We use the ID feature from arcface \cite{deng2018arcface}, a model pre-trained on face recognition tasks, to represent the face identity. Then ID-similarity is measured by the cosine distance of ID features between the input face and the face cropped from the edited image. For each editing prompt and face identity, four images are generated.

\myparagraph{Implementation Details.} 
We choose Stable Diffusion 2.1-base as our base text-to-image model. The learning rate and batch size are set to $5e-5$ and $64$. The encoder is trained for 60,000 iterations. The embedding regularization weight $\lambda$ is set to $1e-4$. Our experiments are trained on a server with eight A100-80G GPUs, which takes about 1 day to complete each experiment. During inference, we use the DDIM \cite{song2020denoising} sampler with 30 steps. The guidance scale is set to 7.5.

\begin{figure}[t]
  \centering
  \includegraphics[width=\linewidth]{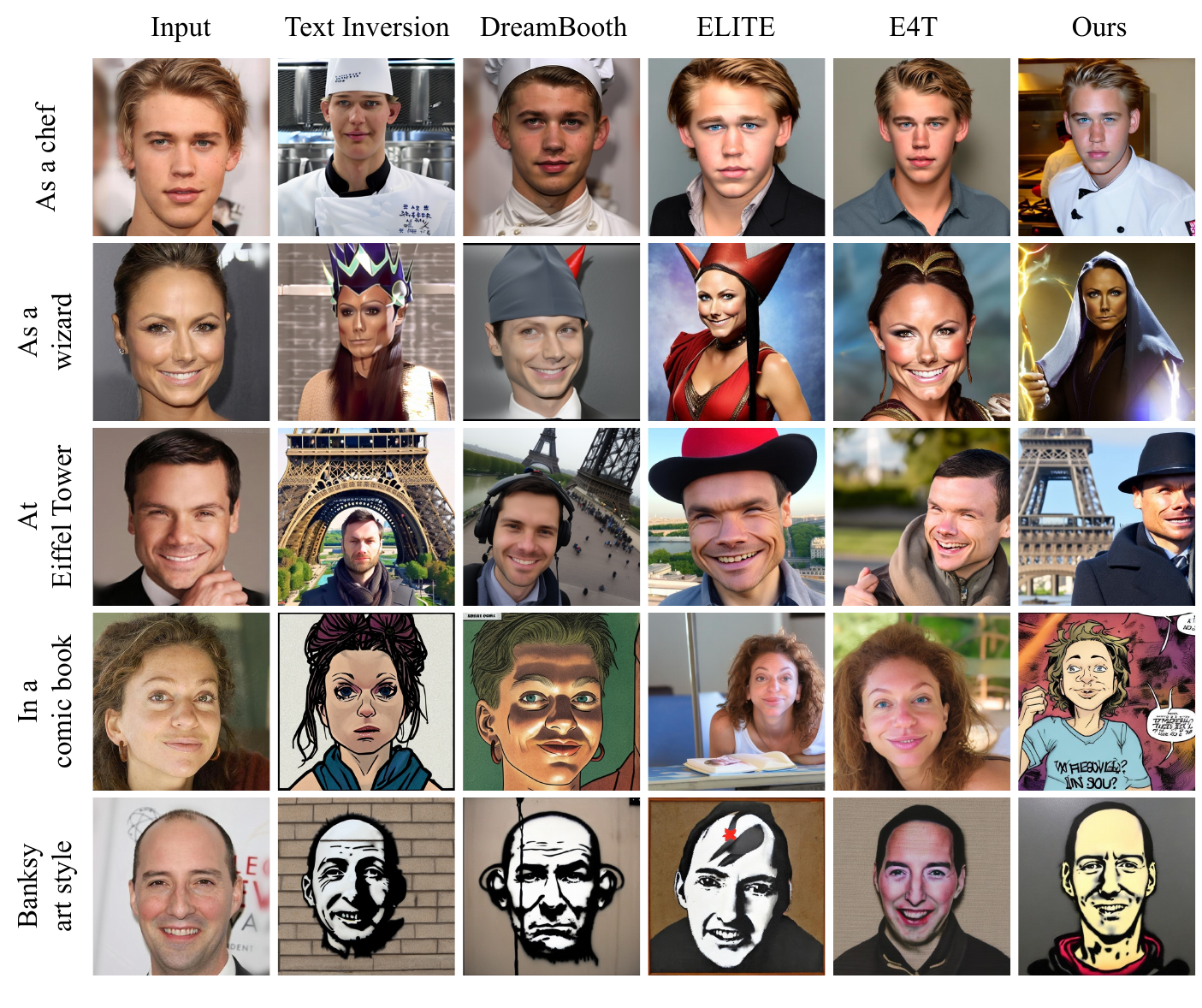}
  \caption{Qualitative comparisons with state-of-the-art methods. \ours \  can generate better text-aligned and ID-preserved images.}
  \label{fig:main_result}
\end{figure}
\begin{table}[t]
\centering
  \caption{Ablation study on $M^2$ ID Encoder. ID encoder with multi-scale feature (MS Feat) and multiple word embeddings (Multi Embedding) achieves best Face-similarity while maintaining a comparable result  Text-alignment metric.}
  \label{tab:id_feat_ablation}
  \begin{tabular}{ccccc}
    \toprule
    ID Encoder & MS Feat & Multi Embedding & Text-alignment $\uparrow$ & Face-similarity $\uparrow$  \\
    \midrule
               &          &             &    \textbf{0.229}   &   0.266   \\
     \checkmark &          &             &  0.228 & 0.302 \\
     \checkmark &  \checkmark &       & \textbf{0.229} & 0.412 \\
      \checkmark &  \checkmark &  \checkmark     & 0.228 & \textbf{0.467} \\
    \bottomrule
  \end{tabular}
\end{table}

\begin{figure}[t]
  \centering
  \includegraphics[width=\linewidth]{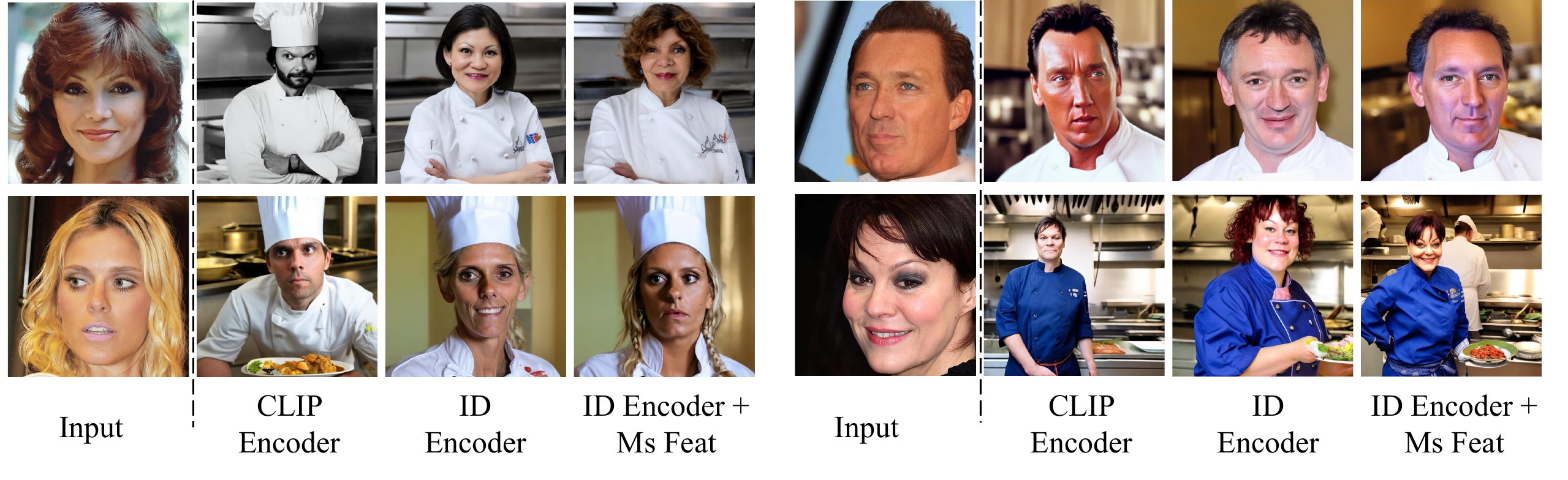}
  \caption{Qualitative comparisons between ID Encoder and the multi-scale features. The editing prompt is "S* as a chef, looking at the camera". We could conclude that both ID Encoder and the multi-scale features greatly improve the ID preservation (\ie, face-similarity).}
  \label{fig:id_encoder}
  \vspace{-2mm}
\end{figure}
\subsection{Comparison to SOTA Methods}
In this section, we compare our method with fine-tuning based methods: Textual Inversion \cite{gal2022image}, DreamBooth \cite{ruiz2022dreambooth} and concurrent works on efficient personalized model: E4T \cite{gal2023designing} which requires finetuning for around 15 iterations for each face, and ELITE \cite{wei2023elite}, a fine-tuning free work. We adopt the widely-used open-sourced Diffusers codebase for Textual Inversion, DreamBooth, and re-implemented  E4T and  ELITE. To ensure a fair comparison, all experiments are conducted with a single face image input. 

\myparagraph{Quantitative and Qualitative Results.} As demonstrated in Tab.\ref{tab:main_result}, our work \ours \ outperforms recent methods across all the metrics, demonstrating superior performance in terms of \editb, \Imetric, and encoding speed. 
We show that \ours \ improves the text-alignment by $7\%$ compared to the second-best E4T \cite{gal2023designing}. Meanwhile,  \ours \ surpasses the second-best model \cite{wei2023elite} on \Imetric \  by $3.7\%$, while enjoying better editability. Benefiting from the direct encoding rather than optimization for unique embeddings, the additional computation cost is only 0.04 s, which can be negligible compared to the time cost (seconds-level) for a standard diffusion-based text-to-image process. The conclusion is further validated by the qualitative results in Fig.\ref{fig:main_result}.

\begin{figure}[t]
\CenterFloatBoxes
\begin{floatrow}
\ttabbox
{\begin{tabular}{cccc}
    \toprule
     {\scriptsize Recon} & {\scriptsize self-aug} & {\scriptsize Text-alignment $\uparrow$} & {\scriptsize Face similarity $\uparrow$ }  \\
    \midrule
    \checkmark &             &  0.213 & 0.380 \\
      &  \checkmark &   0.216 & 0.348 \\
      \checkmark &  \checkmark & \textbf{0.228} & \textbf{0.467} \\
    \bottomrule
  \end{tabular}
  }
  {\caption{Ablation study on self-augmented \editb \  learning. Recon denotes reconstruction training. self-aug denotes self-augmented \editb \  learning, the \editb \ gets improved after applying self-aug.}
  \label{tab:gen_data_ablation}
  }
\killfloatstyle
\ttabbox
{\begin{tabular}{ccc}
    \toprule
     {\scriptsize Emb Num} & {\scriptsize Text-alignment $\uparrow$} & {\scriptsize Face similarity $\uparrow$}  \\
    \midrule
    1           &    \textbf{0.229}   &   0.412   \\
     2           &  0.228 & \textbf{0.462} \\
     3    & 0.188 & 0.472 \\ 
    \bottomrule
  \end{tabular}
  }
  {\caption{Ablation study on the number of word embeddings (Emb Num). Single word embedding could limit the face-similarity while excessive ones may hinder text-alignment.}
  \label{tab:multi_token_ablation}
  }

\end{floatrow}
\end{figure}
\begin{figure}[t]
  \centering
  \includegraphics[width=\linewidth]{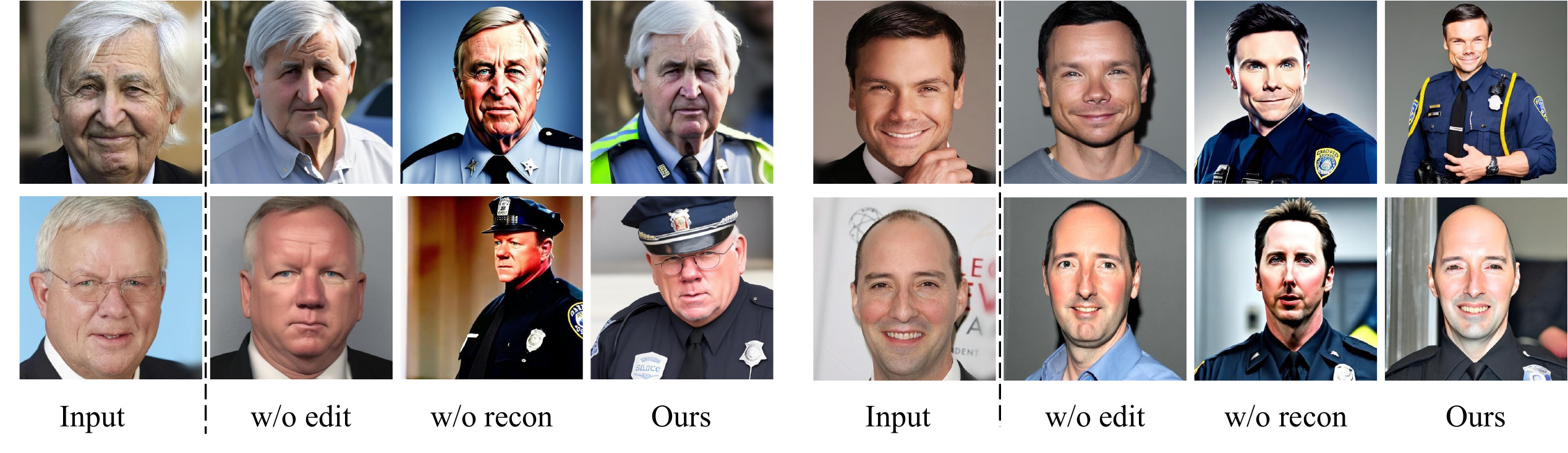}
  \caption{Qualitative comparisons on the self-augmented dataset for \editb \  learning. The editing prompt is "S* as a police, looking at the camera". "w/o edit" and "w/o recon" denote for the encoder is trained without \editb \ learning objective and without reconstruction learning, respectively. We show that the generated images can not follow the prompt properly without the \editb \ learning. Meanwhile, the face similarity will be lower without the reconstruction learning on FFHQ.}
  \label{fig:exp_self_aug}
\end{figure}
\begin{figure}[htb]
  \centering
  \includegraphics[width=\linewidth]{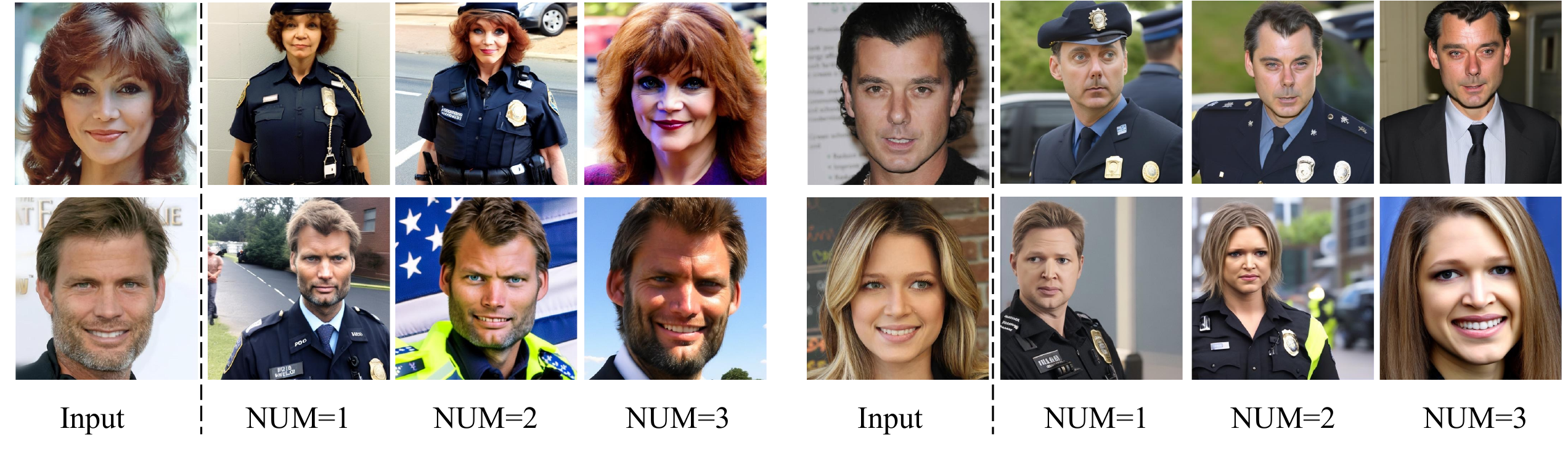}
  \caption{Qualitative comparisons of multiple word embeddings. The editing prompt is "S* as a police, looking at the camera", and "NUM" denotes the number of embeddings.}
  \label{fig:multi_token}
\end{figure}

\subsection{Ablation Studies}
\vspace{-0.2cm}
In this section, we conduct ablation studies to verify the effectiveness of our proposed  $M^2$ ID feature and self-augmented editability learning.
 
\myparagraph{$M^{2}$ ID encoder.} 
We adopt CLIP encoder as our baseline, which is commonly used by concurrent encoder-based methods. Following \cite{wei2023elite, shi2023instantbooth}, we use the last layer CLS feature from CLIP encoder to predict a word embedding. As Fig.\ref{fig:id_encoder} shows, this baseline generally failed to capture the core identity information in the input image, and in some cases, it doesn't even capture the gender information.  
Upon switching from the CLIP encoder to the face-specific ID encoder, the \Imetric \ is improved from $0.266$ to $0.302$, as shown in Tab.\ref{tab:id_feat_ablation}. Integrating the multi-scale features further boosts the \Imetric \ to $0.412$.
Multi-word embeddings are  further utilized to enhance ID-preservation. As shown in Tab.\ref{tab:multi_token_ablation} and Fig.\ref{fig:multi_token}, when we increase the number of embedding  to 2, the Face-similarity is improved by $12\%$ with marginal change of $0.4\%$ on text-alignment. However, when we further increase the number of word embedding,  text-alignment is dropped by $17\%$. We argue that excessive word embeddings may include more information beyond the ID feature such that hinder the editability. Therefore, we choose the embedding number as $2$ to avoid degraded editability.  

\myparagraph{Self-Augmented Editability Learning.} Next, we study the effectiveness of self-augmented editability learning. Fig.\ref{fig:exp_self_aug} indicates that if the model is only trained under the reconstruction objective, the editability \ of embeddings will be limited. To be specific, the model trained without the editability learning objective fails to edit the input identity to a police. Besides, if we only use the limited generated editing dataset, face similarity will be degraded in that there are only around 1000 face IDs in the self-augmented dataset. Combining the reconstruction data (i.e., FFHQ) and generated self-augmented dataset is a better choice to preserve face similarity while following the textual instruction. The quantitative results in Tab.\ref{tab:gen_data_ablation} further confirm our conclusion.

\subsection{Application}
\begin{figure}[H]
  \centering
  \includegraphics[width=\linewidth]{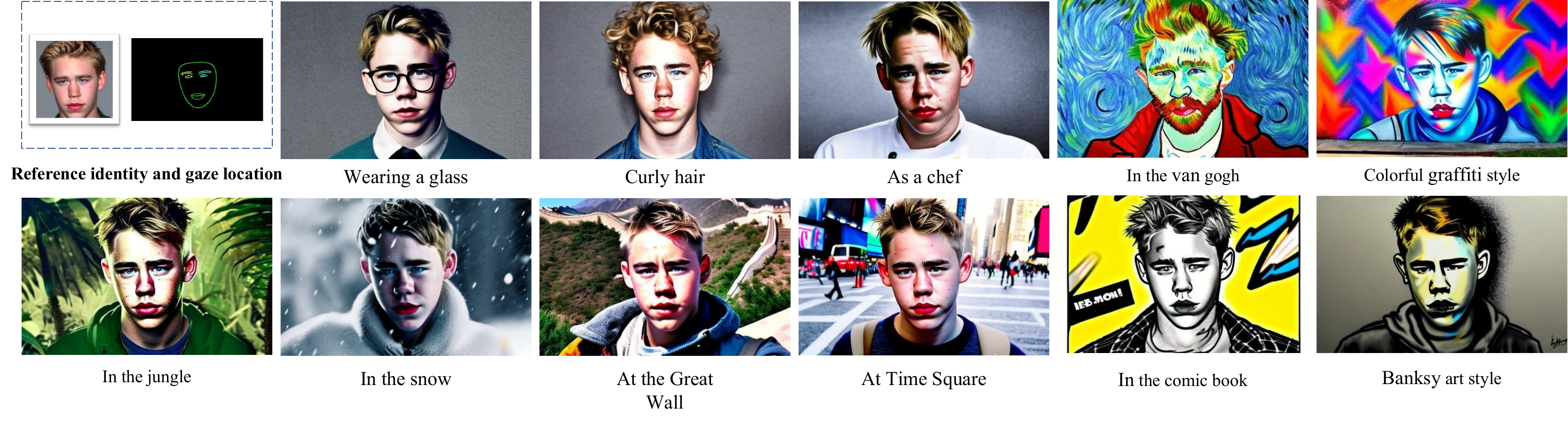}
  \caption{Given a face identify and its gaze location on the canvas, our method can generate a series of images that maintain the same identity while following the editing prompts in the same location.}
  \label{fig:anything}
\end{figure}

\myparagraph{ID-preserved Scene Switch.} As illustrated in Fig.\ref{fig:anything}, given the input face ID and its location in the canvas indicated by the gaze location, we can generate a series of different scene images which share the same identity information and head location with the help of ControlNet \cite{zhang2023adding}. The scene is specified by the text description and can encompass different accessories, hair style, backgrounds, and styles. With this method, we may achieve the effect of "everything and everywhere all at once". 

\section{Limitation}
\setParDis
While our method offers an efficient approach to recreate a human image given one face image, there are several limitations should be noticed. (1) Our model is trained on the high-quality realistic face image dataset, so when the input is a poor-quality face or out-of-domain image, such as a partially obstructed image, the edited image quality is often limited. (2) The \editb \  is undermined when we ask the model to generate a novel scene that may not be satiable for the gender.
\setParDef
\section{Conclusion}\label{sec:con}
\setParDis
In this paper, we present an efficient approach for generating a specified person in new scenes with only one her/his facial image.
The novel $M^2$ ID encoder is proposed to project the identity into multiple word embeddings with multi-scale ID-aware features for the accurate representation of the human in one fast-forward pass with negligible time costs. Besides, the self-augmented editability learning mechanism endows the T2I model with the ability to achieve high editability.
Extensive quantitative and qualitative experiments demonstrate the effectiveness of the proposed methods.
\setParDef

\bibliographystyle{plain}
\bibliography{ref}
\clearpage

\section*{Supplementary}

In this supplementary file, in Section.\ref{Self-Aug}, we will provide the details of constructing the self-augmented dataset. 
In Section.\ref{InstructPix2Pix}, we will compare our method with the recently proposed general editing method InstructPix2Pix \cite{brooks2022instructpix2pix}. In Section.\ref{scene}, we will compare our method with InstructPix2Pix and the baseline that doesn't use identity information on the scene switch application.

\begin{figure}[htb]
  \centering
  \includegraphics[width=\linewidth]{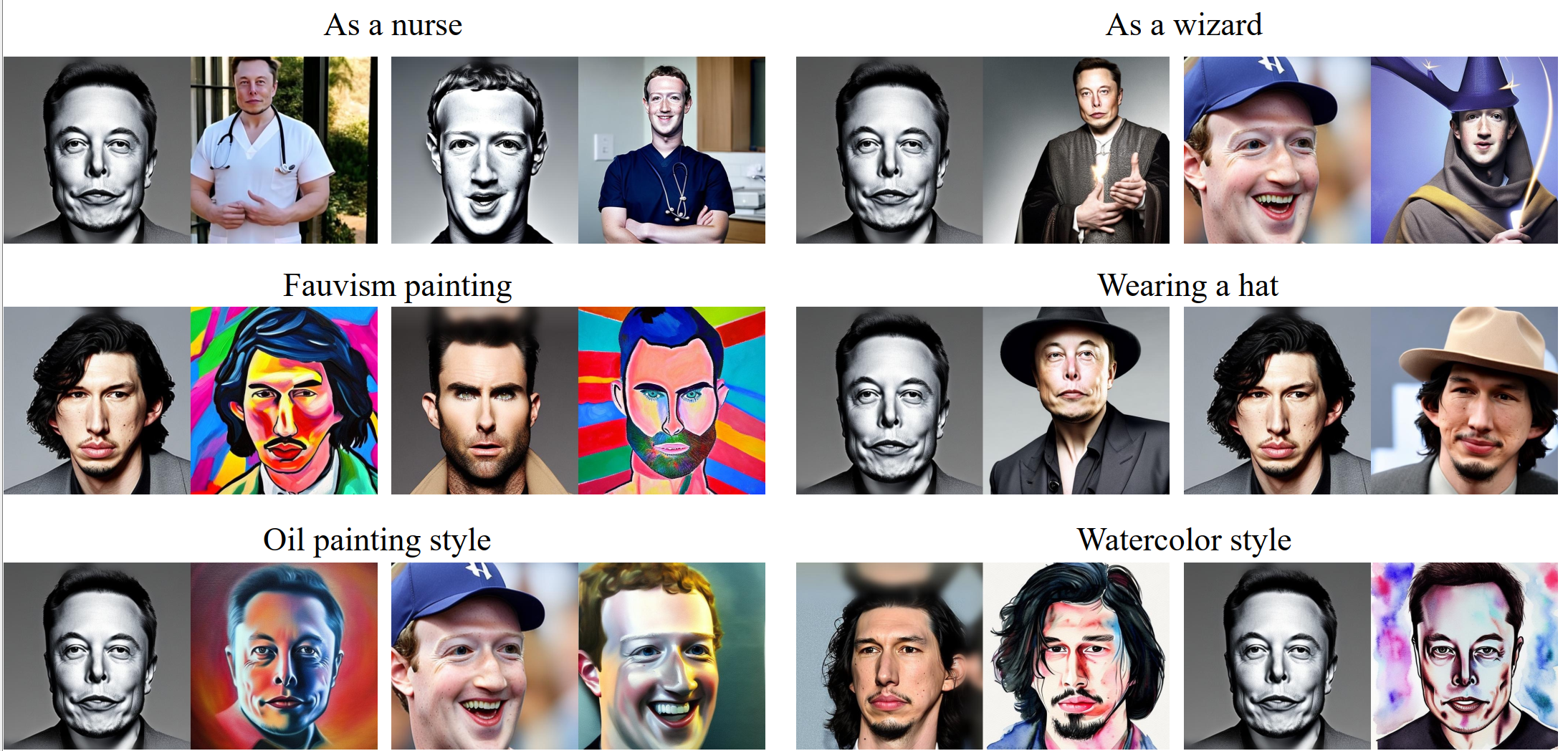}
  \caption{Self-augmented dataset }
  \label{fig:self_aug}
\end{figure}

\begin{figure}[htb]
  \centering
  \includegraphics[width=\linewidth]{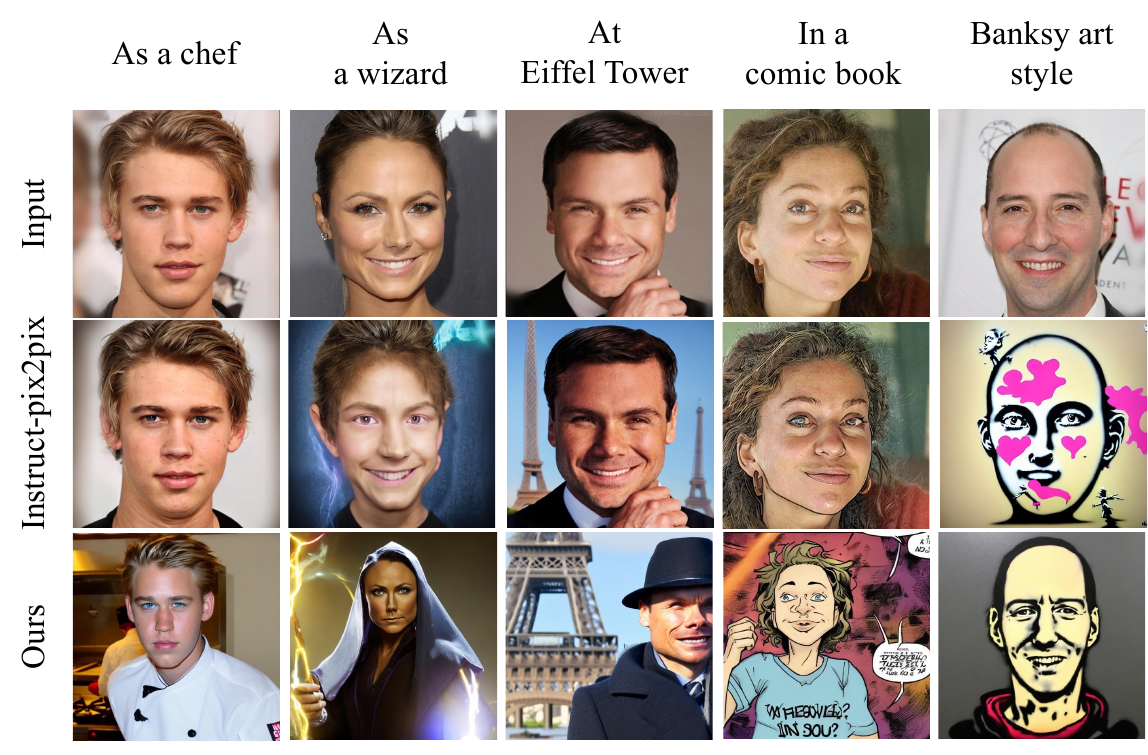}
  \caption{Qualitative comparisons with InstructPix2Pix\cite{brooks2022instructpix2pix}.}
  \label{fig:instruct-p2p}
\end{figure}
\begin{figure}[htb]
  \centering
  \includegraphics[width=\linewidth]{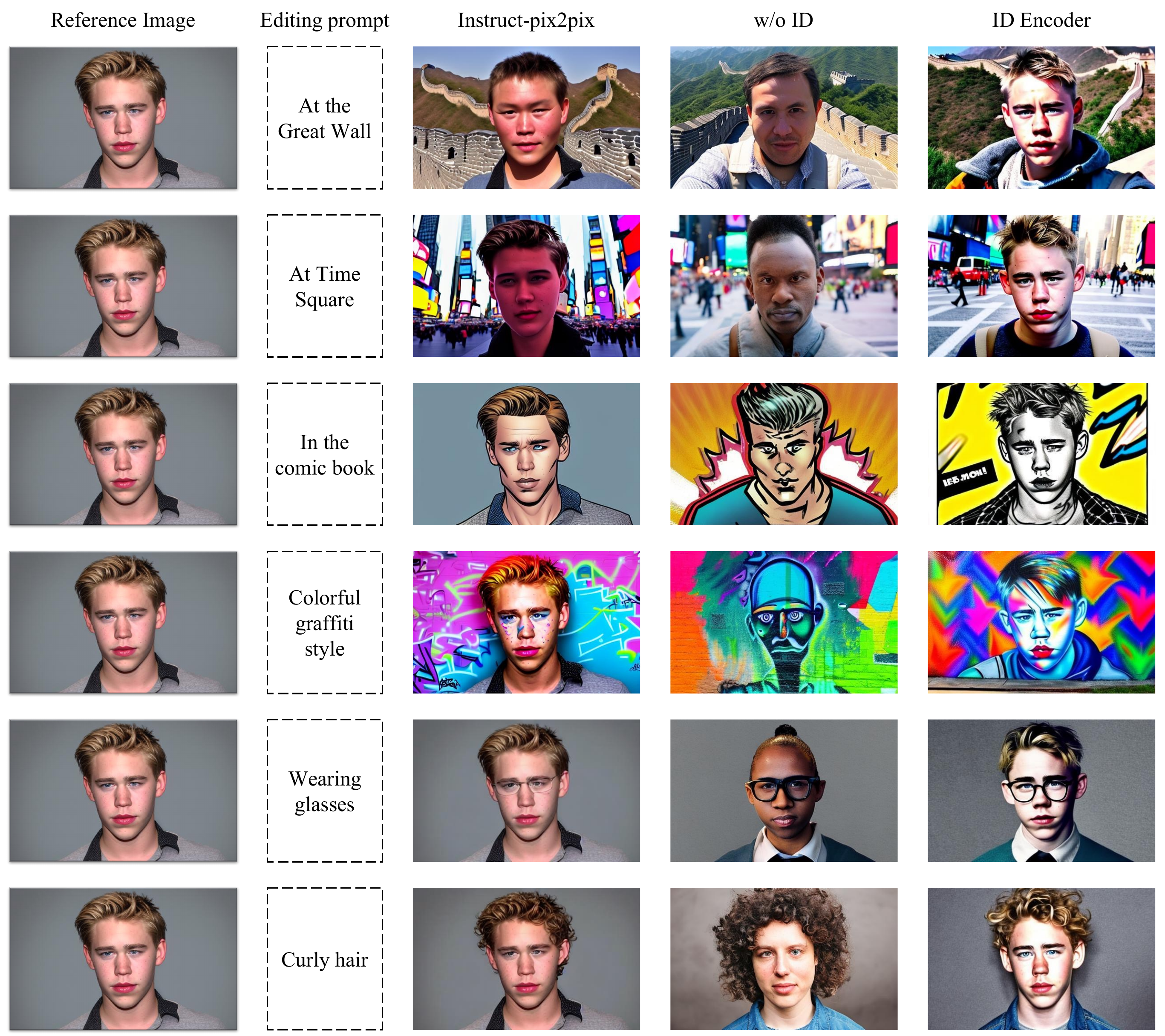}
  \caption{Qualitative comparisons with InstructPix2Pix and w/o ID (without identity information, achieved by replacing $S^{*}$ with "a person"). }
  \label{fig:anything_ablation}
\end{figure}
\appendix
\renewcommand\thesection{\Alph{section}}

\section{Self-Augmented Dataset}\label{Self-Aug}
\myparagraph {Editing Prompts. } The editing prompt list:
\begin{itemize}
\item  Oil painting style, S* face
\item  Watercolor style, S* face
\item  Pencil art style, S* face
\item  Fauvism painting, S* face
\item  S* as a wizard, looking at the camera
\item  S* as a wizard, looking at the camera
\item  S* wearing a hat, looking at the camera
\item  S* as a chef, looking at the camera
\item  S* as a nurse, looking at the camera
\end{itemize}

\myparagraph {Celebrity List. } The celebrity list is in the additional supplementary file, celebrity\_list.txt 

\myparagraph {Training examples. } 
We show the representative training samples in Figure.\ref{fig:self_aug}. 


\section{Qualitative comparisons with InstructPix2Pix\cite{brooks2022instructpix2pix}} \label{InstructPix2Pix}
The results is demonstrated in Figure.\ref{fig:instruct-p2p}. In general, InstructPix2Pix faces challenges when the editing
prompt requries modification of the original image's layout.

\section{Scene Switch}\label{scene}

As depicted in Figure.\ref{fig:anything_ablation}, InstructPix2Pix\cite{brooks2022instructpix2pix} struggles to keep the original identity information in some editing prompts (\eg, "At the Great Wall"). When we only use gaze information, the output images fail to reflect the reference image identity. After adopting our ID encoder to provide ID information, the generated outputs show better identity similarity.

\end{document}